\documentclass{article}

\PassOptionsToPackage{numbers, compress}{natbib}

\usepackage[final]{neurips_2024_ml4ps}

\usepackage[utf8]{inputenc} 
\usepackage[T1]{fontenc}    
\usepackage{hyperref}       
\usepackage{url}            
\usepackage{booktabs}       
\usepackage{amsfonts}       
\usepackage{nicefrac}       
\usepackage{microtype}      
\usepackage{xcolor}         
\usepackage{amsmath}
\usepackage{graphicx}
\usepackage{subcaption}
\usepackage{multirow}
\usepackage{enumitem}

\title{Unpaired Translation of Point Clouds for Modeling Detector Response}

\author{%
Mingyang Li$^{1}$ \quad Michelle Kuchera$^{1}$\quad Raghuram Ramanujan$^{1}$  \\
\textbf{Adam Anthony}$^2$ \quad \textbf{Curtis Hunt}$^3$ \quad \textbf{Yassid Ayyad}$^4$ \\
$^1$ Davidson College \quad $^2$ High Point University \\
\quad $^3$ Michigan State University \quad $^4$ University of Santiago de Compostela\\
\texttt{\{mili, mikuchera, raramanujan\}@davidson.edu}\\
\texttt{\{aanthon2\}@highpoint.edu}\\
\texttt{\{huntc\}@frib.msu.edu}\\
\texttt{\{yassid.ayyad\}@usc.es}\\
}

\begin{document}

\maketitle

\begin{abstract}
Modeling detector response is a key challenge in time projection chambers. We cast this problem as an unpaired point cloud translation task, between data collected from simulations and from experimental runs. Effective translation can assist with both noise rejection and the construction of high-fidelity simulators. Building on recent work in diffusion probabilistic models, we present a novel framework for performing this mapping. We demonstrate the success of our approach in both synthetic domains and in data sourced from the Active-Target Time Projection Chamber.
\end{abstract}

\section{Introduction}
 
Time projection chambers (TPCs) are used for full 3-D reconstruction of events in nuclear and particle physics experiments. The high angular coverage and sensitivity of TPCs allow for the possibility of capturing and analyzing nearly all reaction particles within the chamber. This necessarily increases the complexity of the device, and often leads to detector signals with both electronic and physical noise effects that are difficult to model accurately, especially in the case of gas-filled TPCs, which we study in this paper. Recently, there has been a surge of interest in using machine learning, and particularly generative models, to create theory-agnostic, data-driven simulators in nuclear and particle physics \cite{GenLHC_Butter,ijcai2021p588,MLeventGen, HASHEMI2024100092, ahmad2024comprehensiveevaluationgenerativemodels}.

Despite the promise of generative models, their effectiveness hinges on the nature of the data used for training, presenting a key trade-off. Models trained on experimental data can capture the marginal distribution of the observations, including nuances such as the detector response. However, such data requires unfolding techniques to recover true particle information, limiting their utility in targeted analyses on simulated data. Conditional generative models trained on labeled simulated data offer precise control over the sampling process. But this comes at the cost of simulation fidelity, as such models cannot accurately capture the detector response. Addressing this tension---between the accuracy of the model trained on the experimental data and the controllability of the model trained on the simulations---remains a key problem.

In this work, we address this challenge by presenting a framework for \emph{unpaired translation of point clouds}: given a pair of datasets $\mathcal{D}_X$ and $\mathcal{D}_Y$ comprising point clouds, we train models $G: \mathcal{D}_X \mapsto \mathcal{D}_Y$ and $H: \mathcal{D}_Y \mapsto \mathcal{D}_X$ that can transform events from one domain into events that resemble members of the other domain. In the scenario where $\mathcal{D}_X$ is simulated data and $\mathcal{D}_Y$ is experimental data, $G$ models the detector response for simulated events, while $H$ subtracts noise from experimental events so that they resemble the idealized versions generated by simulators. Crucially, no one-to-one correspondence between the elements of $\mathcal{D}_X$ and $\mathcal{D}_Y$ is required to learn these translational mappings. We evaluate our approach on synthetic point clouds, as well as data sourced from an experiment conducted using the Active-Target Time Projection Chamber (AT-TPC) \cite{BRADT201765} at the National Superconducting Cyclotron Laboratory (NSCL).

\section{Background and Related Work}
\textbf{Diffusion Models} Diffusion probabilistic models (DPMs) have recently gained significant attention due to their impressive results in generative tasks, particularly in areas such as image \cite{nichol2022glide, rombach2021highresolution, hodenoising} and text generation \cite{li2022diffusion,sedd}. A stochastic diffusion process is a Markov chain defined by a series of noise-addition steps that gradually distort the data. The central idea behind diffusion models is to use another Markov chain to convert a sample drawn from a Gaussian distribution back into a sample from the original distribution, by applying learned transformations. In particle physics, there have been significant efforts in using diffusion models for simulation of point cloud data 
\cite{PCJeDi,Buhmann_2023, Buhmann_2024,  PhysRevD.109.012010, ToralesAcosta_2024, buhmann2023epicly, jiang2024choose, quétant2024pipping}. 
Additionally, image-based diffusion models were used for simulation in LArTPCs \cite{PhysRevD.109.072011, PhysRevD.109.012010}. In this work, we adapt a diffusion-based approach proposed by \citeauthor{Luo_2021_CVPR} for generating point clouds, and use it as a building-block in our system \cite{Luo_2021_CVPR}.

\textbf{Unpaired Translation} The task of image translation using conditional GANs was first proposed and studied in the seminal Pix2Pix paper \cite{pix2pix}. This work assumed access to paired data, where each image in the source domain was aligned with a corresponding, translated image in the target domain -- for example, the same scene as viewed during the day and at night. To address the challenge of collecting such data at scale, \citeauthor{cyclegan} employed a novel cycle consistency loss to perform image-to-image translation in an unpaired setting \cite{cyclegan}. Follow-up work has improved upon unpaired translation performance on images using both GANs \cite{zhu2020unpaired, Royer2020} and diffusion models \cite{sasaki2021unitddpm, Zhao2022EGSDEUI}. More recently, \citeauthor{cyclediffusion} proposed CycleDiffusion for unpaired image translation with diffusion models \cite{cyclediffusion}. Their key contribution was the DPM-Encoder, a novel method for encoding input images into a latent space that permitted perfect reconstruction via reverse diffusion. In physics, several applications of unpaired translation have been proposed in recent years \cite{Khan2023, Scherrer:22, Carrazza:2019cnt, torbunov2022, Huang:2023kgs}. Notably, \citeauthor{Huang:2023kgs} used unpaired image translation to map between LArTPC simulations and experiments. Prior unpaired point cloud translation work using diffusion models has focused on the application of 3-D object completion \cite{wen2021c4c,zhang2024reversecomplete}.  \citeauthor{zhengcharacteristic} and \citeauthor{liUnpaired} proposed non-diffusion based unpaired point cloud translation architectures for objects such as furniture \cite{zhengcharacteristic, liUnpaired}. Our work focuses on the unpaired translation of point cloud track data from experimental and simulated AT-TPC data using diffusion models. 

\textbf{AT-TPC} The AT-TPC is a detector that is used to probe the structure of atomic nuclei. In this work, we used data from an AT-TPC experiment studying the fission properties of nuclei in the neutron deficient lead region \cite{Anthony_2023}. A heavy ion beam was produced via $^{208}$Pb fragmentation, and the species of interest separated using the A1900 fragment separator \cite{morrissey2003commissioning, gade2016nscl}. The resulting cocktail beam was identified using the HEIST particle identification system \cite{HEIST} before the beam was directed into the AT-TPC where fission was induced through a fusion-fission reaction using He gas as the target. By collecting the charge produced as the beam and fission fragments travel through the AT-TPC on its highly segmented pad plane, the particle tracks are reconstructed as four-dimensional (three space, one charge) point clouds. The fission experiment produced only two fission products, creating ``Y''-shaped events. This dataset was chosen out of recent AT-TPC experiments as a useful first application because the relative visual simplicity of the events allow for straightforward qualitative evaluation of our results.

\section{Methods}
\label{headings}
We begin with an informal, high-level description of our system which combines CycleDiffusion \cite{cyclediffusion}, an unpaired image translation framework, with a specialized diffusion-based generative model for point clouds \cite{Luo_2021_CVPR}. CycleDiffusion relies on a striking property of diffusion models: that they produce \emph{``uniquely identifiable encodings''} \cite{score}, i.e., diffusion models trained on the same data produce close to identical output images when primed with the same randomly sampled input, regardless of model architecture, or training and sampling procedures \cite{score}. In fact, this alignment of latent representations occurs even when the models are trained on different datasets --- \citeauthor{latent} found that characteristics such as color and texture, as well as high-level semantic features such as pose and gender are preserved when the same latent code is decoded by networks modeling completely different distributions \cite{latent}. This suggests one way to perform unpaired translation between domains: we independently train two diffusion models on the two domains of interest $\mathcal{D}_X$ and $\mathcal{D}_Y$. We encode a source point cloud using its in-domain encoder, but then decode the resultant latent representation with the other diffusion model. This, in effect, is CycleDiffusion \cite{cyclediffusion}. We leverage \citeauthor{Luo_2021_CVPR}'s point cloud diffusion architecture to adapt this idea to the translation of point clouds \cite{Luo_2021_CVPR}. We discuss some of the relevant technical details in the rest of this section.

\subsection{Architecture Details}
\begin{figure}[htbp]
    \centering
    \begin{minipage}[htbp]{0.49\textwidth}
        \centering
        \includegraphics[height=1.5in]{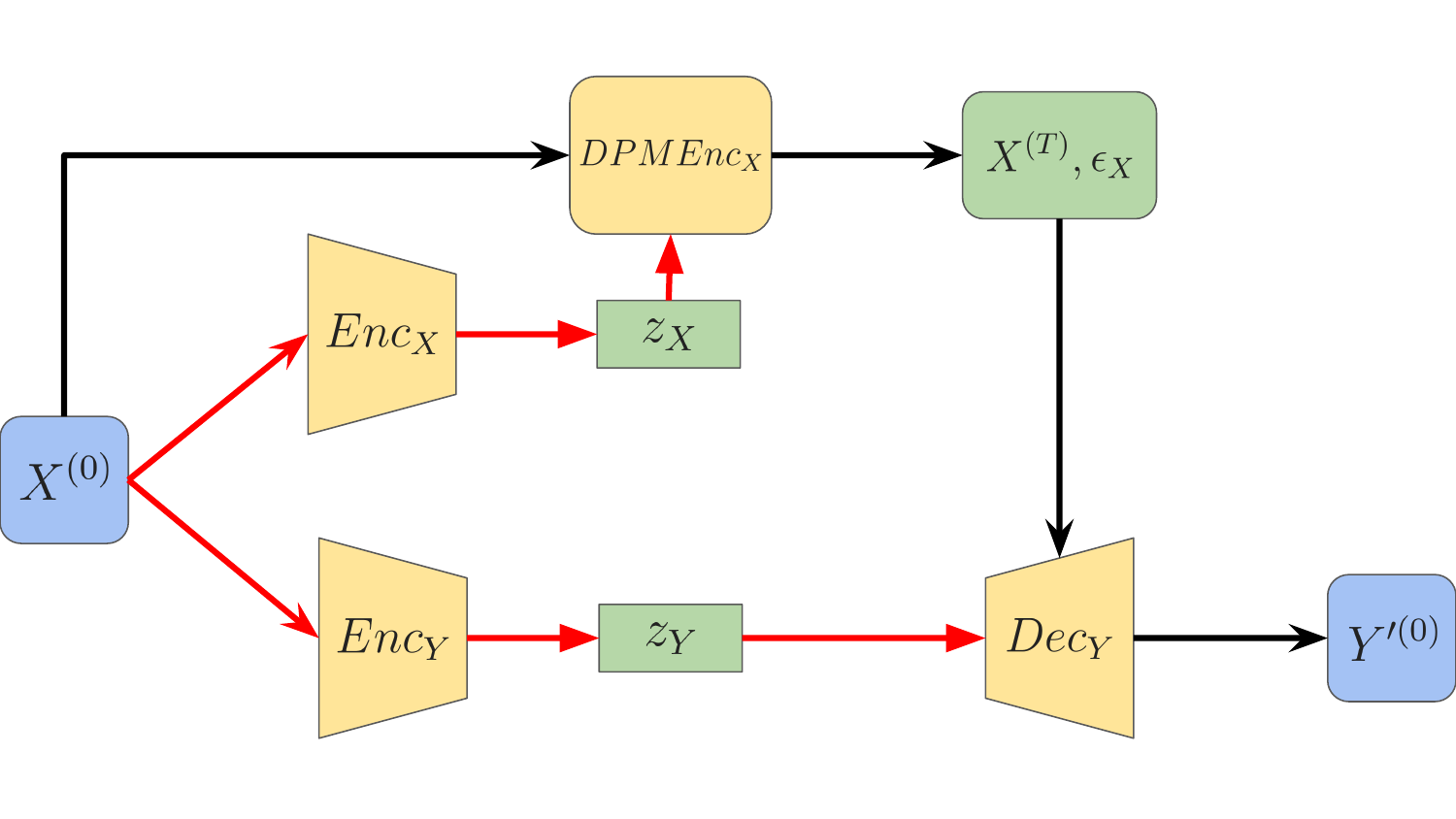}
        \caption{Schematic depicting our unpaired translation model at inference time. Arrows in red denote our modifications to the standard CycleDiffusion architecture to incorporate a point cloud diffusion model.}
        \label{fig:arch}
    \end{minipage}%
    \hfill
    \begin{minipage}[htbp]{0.49\textwidth}
        \centering
        \begin{tabular}{ll}
            \toprule
            \textbf{Hyperparameter} & \textbf{Value} \\ 
            \midrule
            Initial Learning Rate     & 0.001          \\ 
            Final Learning Rate       & 0.0001         \\
            Optimization Algorithm    & Adam \cite{adam} \\
            Batch Size                & 128            \\ 
            Number of Training Iterations      & 1,000,000      \\ 
            Latent Dimension Size     & 256            \\ 
            Diffusion Steps           & 256            \\ 
            \bottomrule
        \end{tabular}
        \captionof{table}{Hyperparameters used during training the DPMs.}
        \label{tab:hyper}
    \end{minipage}
\end{figure}

We begin by first training two independent point cloud DPMs, one for each domain $\mathcal{D}_X$ and $\mathcal{D}_Y$. These models use the encoder-decoder architecture described by \citeauthor{Luo_2021_CVPR}, modified to accept our 4-D point cloud data \cite{Luo_2021_CVPR}. The encoders are standard PointNet encoders \cite{pointnet}, while the decoders are neural networks implementing a reverse diffusion process \cite{Luo_2021_CVPR}. The values of key hyperparameters used for training these models are given in Table~\ref{tab:hyper}. Once training is complete, the components of the DPMs are rewired as shown in the schematic in Figure~\ref{fig:arch}. Here, $Enc_X$ ($Enc_Y$) and $Dec_X$ ($Dec_Y$) denote the trained encoder and decoder for domain $\mathcal{D}_X$ ($\mathcal{D}_Y$) respectively. The $DPMEnc_X$ module is a DPM-Encoder which is described in greater detail in Section~\ref{sec:dpmencoder}. This component contains no trainable parameters. To translate an event $\boldsymbol{X}^{(0)}$ from domain $\mathcal{D}_X$ to domain $\mathcal{D}_Y$, we start by processing the event through both encoders, $Enc_X$ and $Enc_Y$, to obtain its latent representations $\boldsymbol{z}_X$ and $\boldsymbol{z}_Y$ within each respective domain. Subsequently, we pass the latent representation $\boldsymbol{z}_X$ along with the event $\boldsymbol{X}^{(0)}$ to $DPMEnc_X$, which simulates the forward diffusion process for $T$ steps producing the event $\boldsymbol{X}^{(T)}$ along with the estimated noise added at each step $\boldsymbol{\epsilon}_X$. Note that $\boldsymbol{\epsilon}_X = \boldsymbol{\epsilon}_T \oplus \ldots \boldsymbol{\epsilon}_2 \oplus \boldsymbol{\epsilon}_1$, where $\oplus$ denotes concatenation. In the final step, the decoder for the $\mathcal{D}_Y$ domain, $Dec_Y$, uses $\boldsymbol{Y}'^{(T)} = \boldsymbol{X}^{(T)}$ as the initial point cloud and starts the decoding process, guided by $\boldsymbol{z}_Y$ and $\boldsymbol{\epsilon}_X$. One single decoding step is computed as
\begin{equation}
    \label{eq:pythagorean}
    \boldsymbol{Y}'^{(t-1)} = \boldsymbol{\mu}_Y(\boldsymbol{Y}'^{(t)}, t, \boldsymbol{z}_Y) + {\sigma}_t\boldsymbol{\epsilon}_t \qquad \qquad t = T, \ldots, 1
\end{equation}
where $\boldsymbol{\mu}_Y$ is the learned mean estimator in $Dec_Y$ \cite{Luo_2021_CVPR}, which estimates the mean point cloud $p(\boldsymbol{Y^{(t-1)}}|\boldsymbol{Y^{(t)}})$, and stochastically at time step $t$ is added by ${\sigma}_t\boldsymbol{\epsilon}_t$, where $\boldsymbol{\epsilon}_t \sim \mathcal{N}(0,\boldsymbol{I})$ and $\sigma_t$ is the variance of the added noise at time step $t$ which increases linearly with time. The result of the final decoding step $\boldsymbol{Y}'^{(0)}$ is the translation of the original event $\boldsymbol{X}^{(0)}$ into domain $\mathcal{D}_Y$. To perform the translation in the opposite direction, we can simply swap $X$ and $Y$ in the preceding discussion. The code to reproduce our model and results can be accessed at the following repository: \url{https://github.com/alpha-davidson/attpc_unpaired_trans}.

\subsection{Modified DPM-Encoder}
\label{sec:dpmencoder}

\citeauthor{cyclediffusion} proposed the DPM-Encoder as a means to deterministically map images to a latent code in stochastic diffusion models \cite{cyclediffusion}. Their key insight was to retain the noise that was added to the samples during each step of the forward diffusion process, together with the final output $\boldsymbol{X}^{(T)}$, and to treat the concatenation of these quantities as the latent encoding of the original input image. While this is a relatively straightforward idea, it leads to impressive results in a variety of tasks, including unpaired translation of images \cite{cyclediffusion}. Unlike image data, however, point cloud events are permutation invariant, and thus, we cannot use the DPM-Encoder directly in our framework. We augment the DPM-Encoder with an additional shape latent that is computed using a PointNet encoder (see Figure~\ref{fig:arch}). This enables the encoding of topological information about the original point cloud $\boldsymbol{X}^{(0)}$ in the latent representation, prior to decoding and translation via the reverse diffusion process in $Dec_Y$.






\subsection{Data}
We evaluate our approach on three different domains, in each case translating between a ``clean'' and a ``noisy'' point cloud.
\begin{enumerate}[leftmargin=*]
\item $(\mathcal{L}_X, \mathcal{L}_Y)$: a pair of synthetic point cloud datasets comprising clean and noisy line segments, respectively, oriented along the $z$ axis. $\mathcal{L}_X$ is a proxy for simulated data, while $\mathcal{L}_Y$ is a stand-in for experimental data. Each line is composed of points with $x=0$, a fixed $y$ sampled from $U(0, 2)$ and 256 $z$ values sampled from $U(0, 2)$. We corrupt the elements of $\mathcal{L}_Y$ by adding noise sampled from $\mathcal{N}(0, 0.1y)$ to each point. This task is designed to evaluate the model's ability to capture simple non-uniform noise distributions. $\mathcal{L}_X$ and $\mathcal{L}_Y$ each consist of 1,000 events.
\item $(\mathcal{G}_X, \mathcal{G}_Y)$: a pair of synthetic datasets containing clean and noisy geometric shapes (triangular prisms and cuboids), respectively, of different sizes and orientations. The shapes are created by sampling points uniformly at random along the edges of the solid. We add Gaussian noise to each point in each example in $\mathcal{G}_Y$ to create the noisy examples. This task is intended to assess the model's ability to perform translations between two domains in a multi-class setting. $\mathcal{G}_X$ comprises 2,000 examples, evenly distributed between 1,000 triangular prisms and 1,000 cuboids; $\mathcal{G}_Y$ contains noisy versions of the same examples.
\item $(\mathcal{A}_X, \mathcal{A}_Y)$: simulated and experimental fission events from the AT-TPC, respectively. These events have a characteristic ``Y'' shape, where the base of the Y is the incoming beam and the other two legs are outgoing the fission fragments. Our simulated dataset consists of 5,000 events, while the experimental dataset includes 11,500 events selected by applying a cut on the fission angle to filter out events where no fission occurred.
\end{enumerate}

\section{Results}
\label{others}
\begin{figure}[htbp]
    \centering
    \small
    \begin{minipage}{0.45\textwidth}
        \centering
        \includegraphics[width=2in, height=1in]{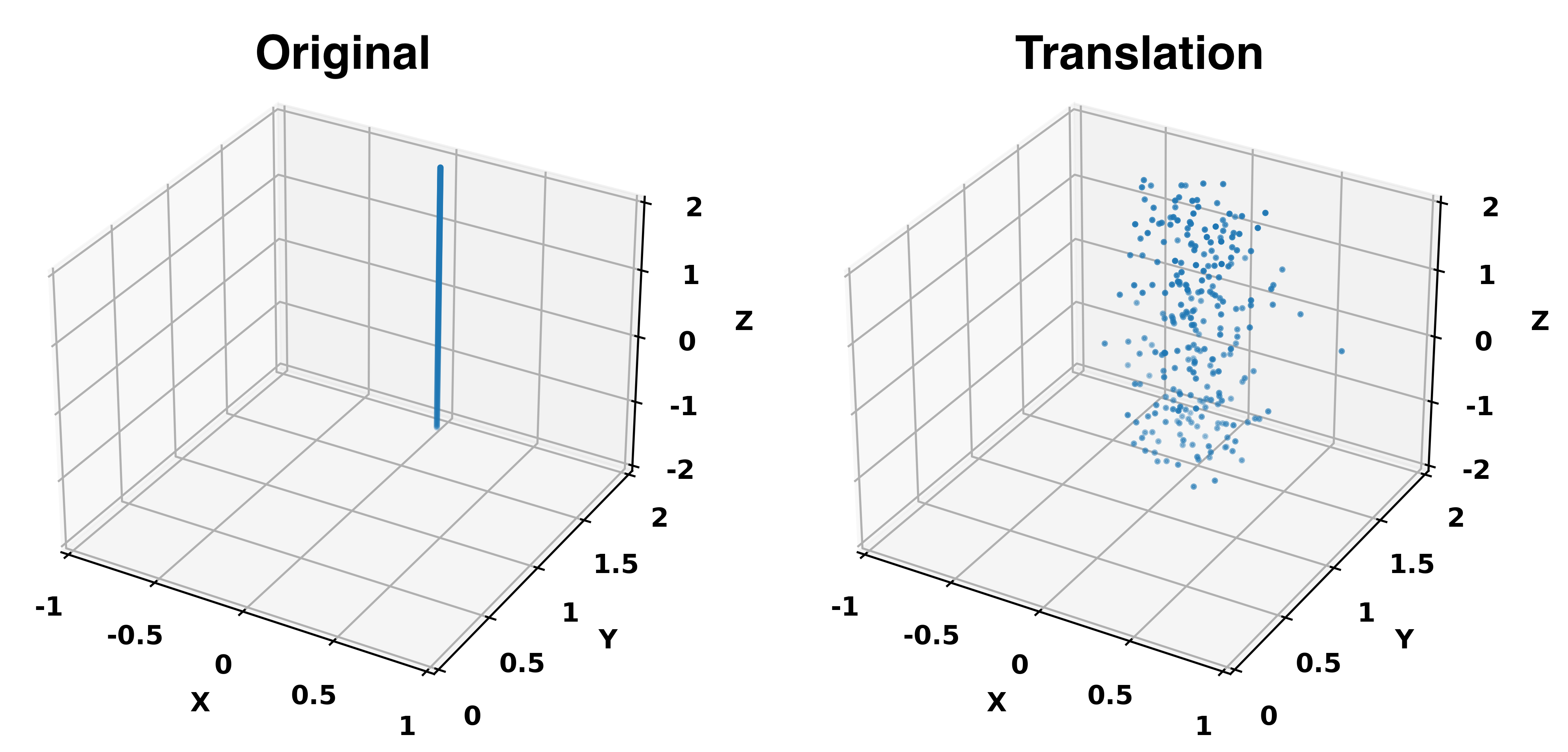}
    \end{minipage}
    \hspace{0.08\textwidth} 
    \begin{minipage}{0.45\textwidth}
        \centering
        \includegraphics[width=2in, height=1in]{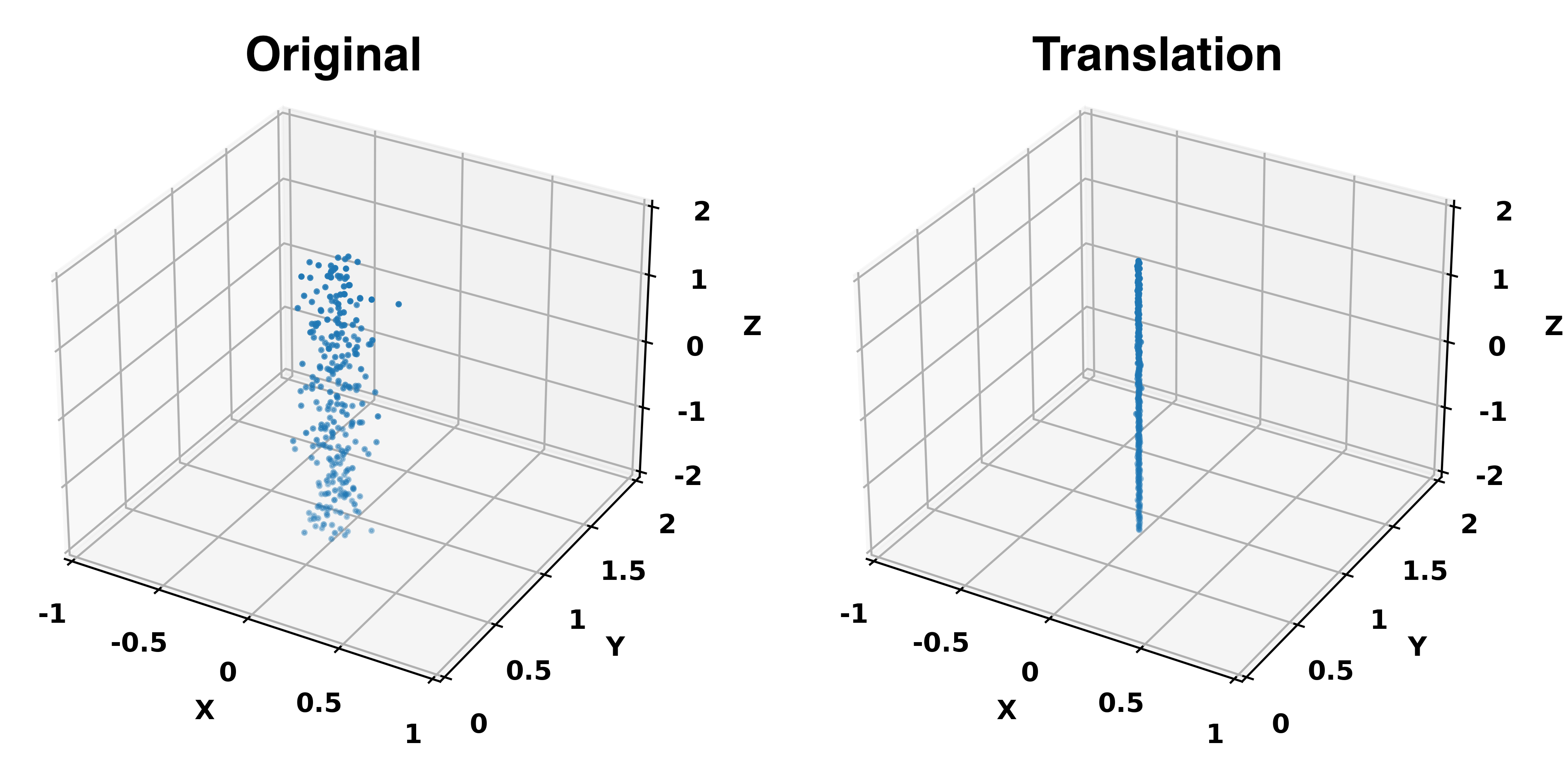}
    \end{minipage}
    \caption{Samples of translation results on 
    on the ($\mathcal{L}_X, \mathcal{L}_Y$) dataset in two directions.}
    \label{fig:line}
\end{figure}

We evaluate the ability of our model to learn a conditional noise distribution in translation on the $(\mathcal{L}_X, \mathcal{L}_Y)$ dataset by calculating the fitted standard deviation and error after translation. The quantitative results can be seen in Tab.~\ref{tab:line}, and qualitative translation results are shown in Fig.~\ref{fig:line}. 

\begin{table}[htbp]
  \centering
  \small
  \begin{tabular}{llll}
    \toprule
    $y$  &  $\sigma(y)$   & $\sigma_T \pm \sigma_T/\sqrt{N}$   & MAE \\ 
    \midrule
    0.04 & .004 & 0.010 $\pm$ 0.0004& 0.005\\
    0.63 & .063 & 0.076 $\pm$ 0.003 & 0.013 \\
    1.32 & .132 & 0.121 $\pm$ 0.005  &0.009 \\
    1.99 & .199 & 0.181 $\pm$ 0.008 &  0.015 \\
    \bottomrule
  \end{tabular}
  \vspace{0.2cm}
  \caption{Standard deviations of noise added by the $\mathcal{L}_X \rightarrow \mathcal{L}_Y$ translation process using our model. We report the ground truth standard deviation $\sigma(y) = 0.1y$, the standard deviation, $\sigma_T$, of translated events with estimated standard error $\sigma_T/\sqrt{N}$, and the mean absolute error between $\sigma(y)$ and $\sigma_T$. The consistently small MAE values demonstrate that the translation model is capable of learning conditional noise relationships.}
  \label{tab:line}
\end{table}

\begin{figure}[htb!]
    \centering
    \small
    \begin{minipage}{0.45\textwidth}
        \centering
        \includegraphics[width=2.4in]{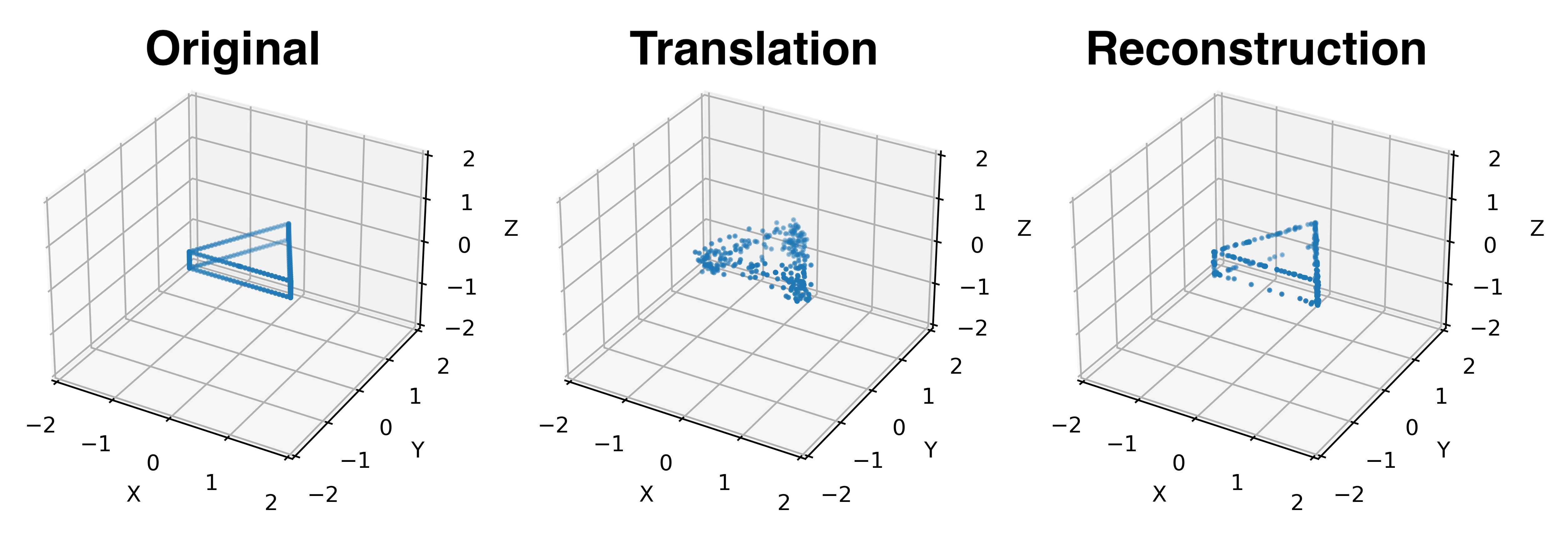}
    \end{minipage}
    \hspace{0.08\textwidth} 
    \begin{minipage}{0.45\textwidth}
        \centering
        \includegraphics[width=2.4in]{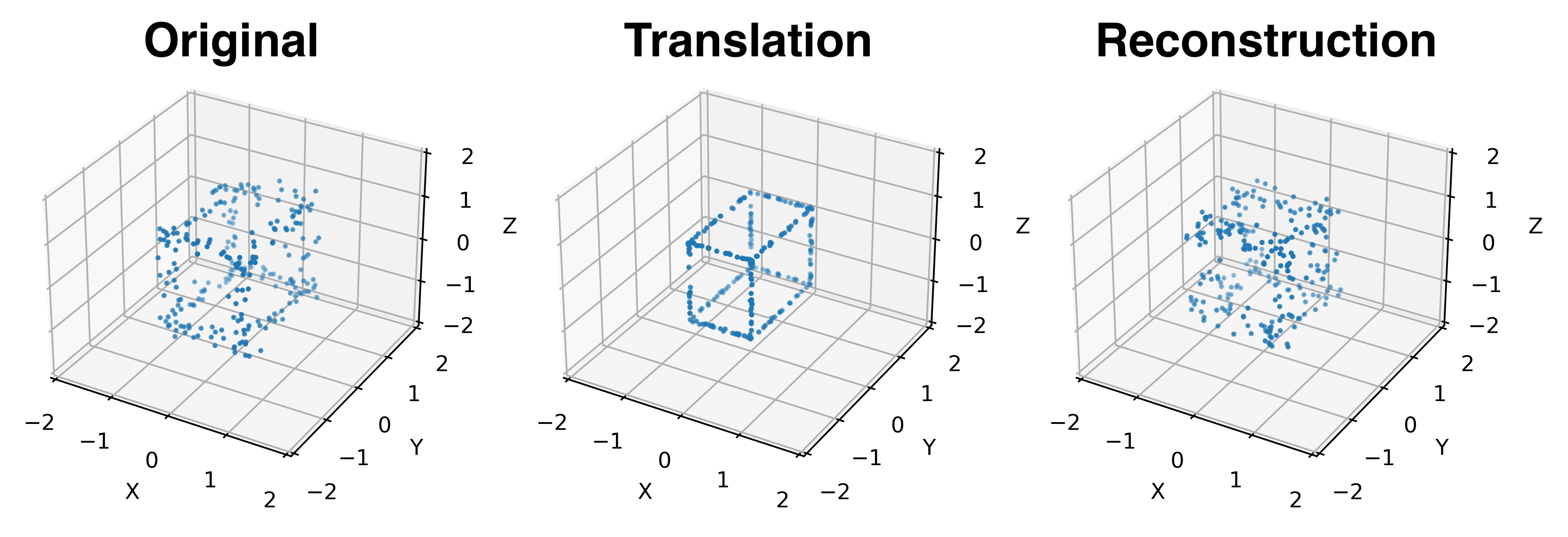}
    \end{minipage}
    \centering
    \small
    \begin{minipage}{0.45\textwidth}
        \centering
        \includegraphics[width=2.6in]{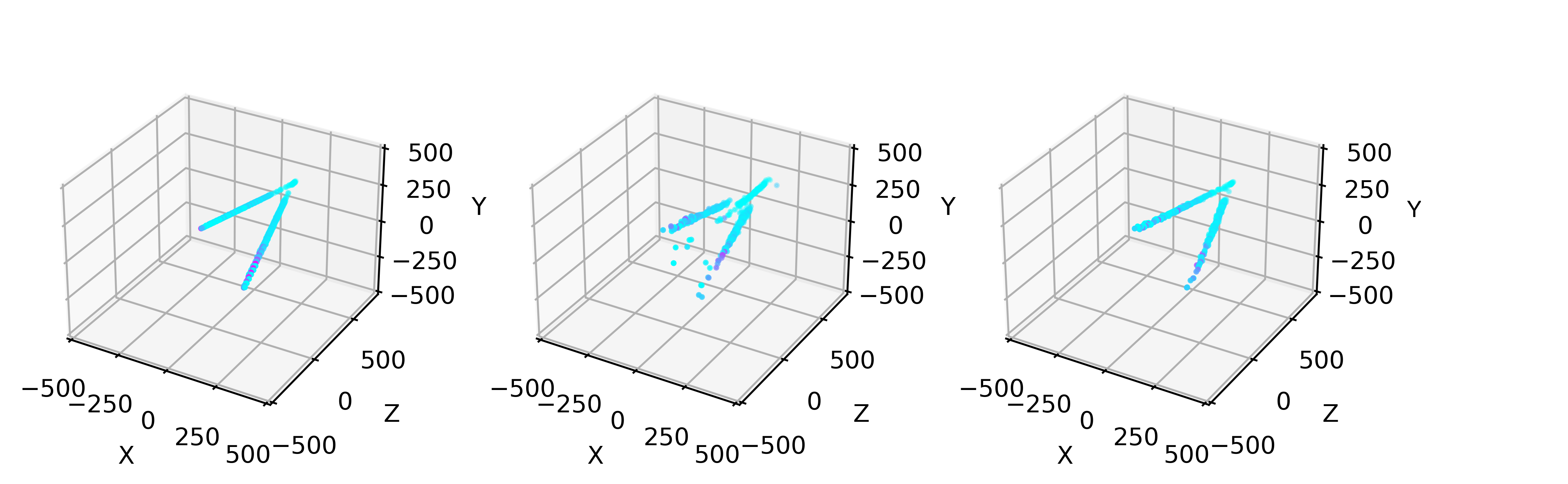}
    \end{minipage}
    \hspace{0.05\textwidth} 
    \begin{minipage}{0.45\textwidth}
        \centering
        \includegraphics[width=2.6in]{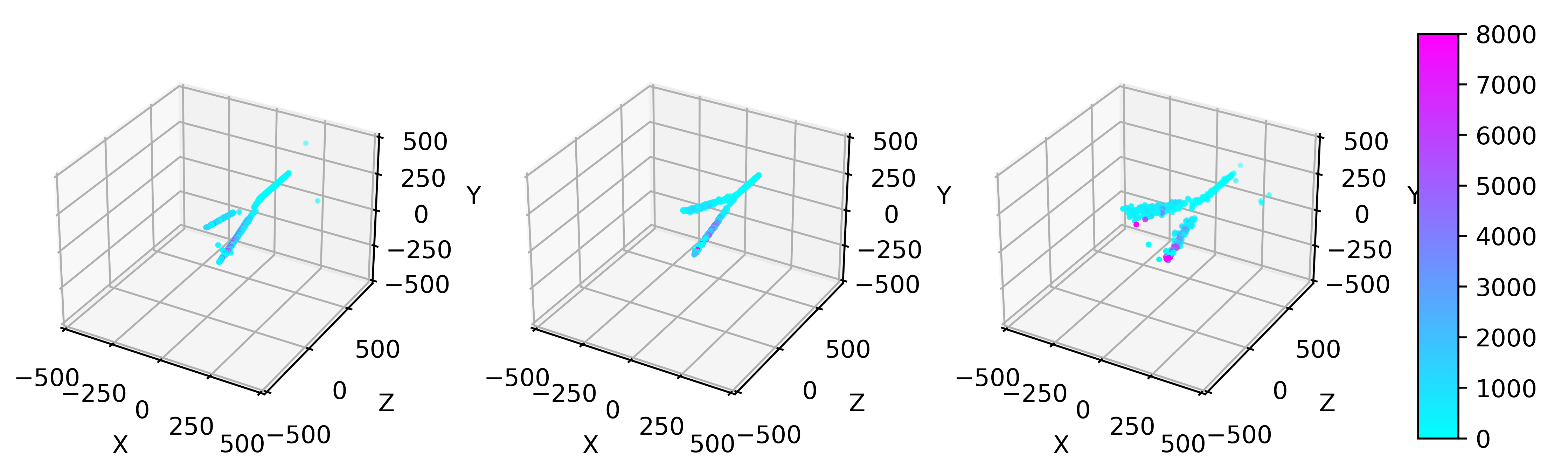}
    \end{minipage}
    \caption{Translation and reconstruction results on $(\mathcal{G}_X, \mathcal{G}_Y)$ (top) and $(\mathcal{A}_X, \mathcal{A}_Y)$ data (bottom).}
    \label{fig:trans2}
\end{figure}

We evaluate the translation and reconstruction quality of our model on the $(\mathcal{G}_X, \mathcal{G}_Y)$ and $(\mathcal{A}_X, \mathcal{A}_Y)$ datasets. Given a domain $D$, we calculate the Jensen-Shannon divergence (JSD) between events translated into domain $D$ and the original events in $D$ to quantify the quality of the translation (JSD(trans)). We compare this to two baseline JSD scores: one computed between different batches of events drawn from within the same domain $D$ (JSD(in-domain)), and one computed between a batch of randomly generated point clouds and events from $D$ (JSD(rand)). We use Chamfer Distance (CD) to measure reconstruction accuracy after performing bidirectional translations (for example, $\mathcal{G}_X \rightarrow \mathcal{G}_Y \rightarrow \mathcal{G}_X$). Quantitative results are shown in Tab.~\ref{tab:res2}, and visual results are shown in Fig.~\ref{fig:trans2}.

\begin{table}[htb!]
  \centering
  \small
  \begin{tabular}{lllllll}
    \toprule
    Model & JSD(trans) & JSD(in-domain) & JSD(rand) & CD(reco) & $\sigma_{CD(reco)}$ & CD(clean)\\
    \midrule
     $\mathcal{G}_X\rightarrow \mathcal{G}_Y$ & 0.323 & 0.314 & 0.711 & 0.163 & 0.179 & --\\
    $\mathcal{G}_Y\rightarrow \mathcal{G}_X$ & 0.382 & 0.270 & 0.701 & 0.153 & 0.157 & --\\
    \midrule
    $\mathcal{A}_X\rightarrow \mathcal{A}_Y$ & 0.044 & 0.005 & 0.701 & 0.053 & 0.066 & --\\
    $\mathcal{A}_Y\rightarrow \mathcal{A}_X$ & 0.005 & 0.004 & 0.703 & 0.269 & 0.523 & 0.079\\
    \midrule
    \bottomrule
  \end{tabular}
  \vspace{0.2cm}
  \caption{Evaluation metrics for the models trained using the $(\mathcal{G}_X, \mathcal{G}_Y)$ and $(\mathcal{A}_X, \mathcal{A}_Y)$ datasets. JSD is used to evaluate translation quality, and is compared with JSDs of in-domain and random data. CD is used to evaluate reconstruction quality. For experimental data reconstruction, we remove the extreme outliers, which comprised $\sim 1\%$ of the reconstructed events, to create a clean reco subdataset. Our results on the  $(\mathcal{G}_X, \mathcal{G}_Y)$ dataset provide a baseline performance, and our comparable results on the  $(\mathcal{A}_X, \mathcal{A}_Y)$ datasets provide support that this approach is suitable for our application, aside from a small percentage of outliers.}
  \label{tab:res2}
\end{table}

\section{Conclusions}
\citeauthor{cyclediffusion} proposed a DPM-Encoder that maps images into an exactly reconstructable latent space. We used DPMs with a modified DPM-Encoder to address the challenge of unpaired translation in AT-TPC events. By effectively translating between simulated and experimental datasets, our model offers an approach for generating realistic detector responses and reducing noise in experimental data. 

\begin{ack}
This material is based on work supported by the National
Science Foundation under grants OAC-2311263
 and PHY-2012865.
\end{ack}

{
\small
\bibliographystyle{unsrtnat}
\bibliography{refs.bib}
}

\end{document}